\def\year{2021}\relax
\documentclass[letterpaper]{article} 
\usepackage{aaai21}  
\usepackage{times}  
\usepackage{helvet} 
\usepackage{courier}  
\usepackage[hyphens]{url}  
\usepackage{graphicx} 
\usepackage{natbib}  
\usepackage{caption} 
\usepackage{subfig}
\usepackage{algorithm}
\usepackage[noend]{algpseudocode}
\usepackage{multirow}
\usepackage{amsmath}

\title{Searching for Alignment in Face Recognition}

\author{Xiaqing Xu\textsuperscript{\rm 1},
		Qiang Meng\textsuperscript{\rm 1},
		Yunxiao Qin\textsuperscript{\rm 2},
		Jianzhu Guo\textsuperscript{\rm {3,4}}, \\
	    Chenxu Zhao\textsuperscript{\rm 5}\thanks{Corresponding Author.},
	    Feng Zhou\textsuperscript{\rm 1},
	    Zhen Lei\textsuperscript{\rm {3,4}} \\
}
\affiliations {
	\textsuperscript{\rm 1}AIBEE, Beijing, China,\
	\textsuperscript{\rm 2}Northwestern Polytechnical University, Xian, China\\
	\textsuperscript{\rm 3} CBSR \& NLPR, Institute of Automation, Chinese Academy of Sciences, Beijing, China \\
	\textsuperscript{\rm 4} School of Artificial Intelligence, University of Chinese Academy of Sciences \\
	\textsuperscript{\rm 5}Academy of Sciences, Mininglamp Technology, Beijing, China \\
    \{xqxu; qmeng; fzhou\}@aibee.com,
	qyxqyx@mail.nwpu.edu.cn,
	\{jianzhu.guo; zlei\}@nlpr.ia.ac.cn
	zhaochenxu@mininglamp.com \\
}
\begin{document}

\maketitle

\begin{abstract}
A standard pipeline of current face recognition frameworks consists of four individual steps: locating a face with a rough bounding box and several fiducial landmarks, aligning the face image using a pre-defined template, extracting representations and comparing.
Among them, face detection, landmark detection and representation learning have long been studied and a lot of works have been proposed.
As an essential step with a significant impact on recognition performance, the alignment step has attracted little attention.
In this paper, we first explore and highlight the effects of different alignment templates on face recognition. Then, for the first time, we try to search for the optimal template automatically. We construct a well-defined searching space by decomposing the template searching into the \textit{crop size} and \textit{vertical shift}, and propose an efficient method \textbf{F}ace \textbf{A}lignment \textbf{P}olicy \textbf{S}earch (FAPS).
Besides, a well-designed benchmark is proposed to evaluate the searched policy.
Experiments on our proposed benchmark validate
the effectiveness of our method to improve face recognition performance.



\end{abstract}

\section{Introduction}

Face recognition is a long-standing topic in the research community of computer vision.
A standard pipeline of the recognition framework consists of four individual steps: locating faces with bounding boxes and  fiducial points, aligning face images using a pre-defined template, extracting face representations and representation comparing.
The second step, also named as face alignment (in Fig. \ref{fig:align}), serves as deforming face images such that  fiducial points are spatially aligned and
simplifies the recognition task by normalizing the in-plane rotation, scale and translation variations. However, most  recent works
\cite{taigman2014deepface,sun2014deep,schroff2015facenet,liu2017sphereface,Deng_2019_CVPR,kang2019attentional} on face recognition focus on designing loss functions and exploring network structures. In contrast, the alignment procedure before model training is less studied.

\begin{figure}
    \centering
    \includegraphics[width=0.42\textwidth]{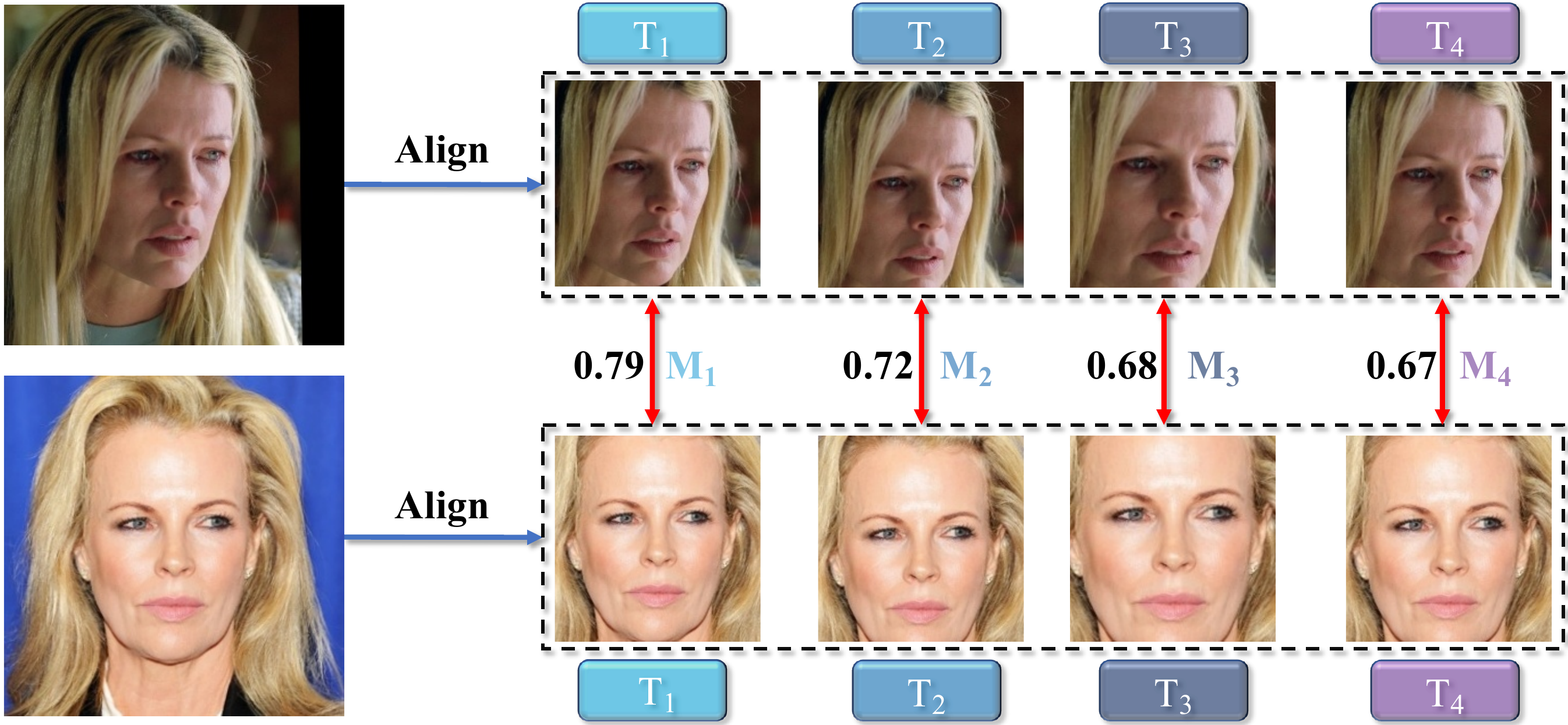}
    \caption{\normalsize{Verification results based on different face templates. Models $M_i, i=1,2,3,4$ are trained with samples aligned by templates $T_i, i=1,2,3,4$ respectively.  Significant differences between cosine similarities are observed.}
    }
    \label{fig:intro_1}
\end{figure}

In this paper, we first explore the effects of the alignment templates\cite{Deng_2019_CVPR,zhu2019large,guo2020learning} on face recognition performance. Face features can be divided into two sets depending on the zone where they are located: internal features, including eyes, nose and mouth, and external features, composed by the hair, chin and face outline. The benefits of external information have been observed in some early works \cite{lapedriza2005external,Andrews2010Internal},
but they are rarely discussed in the modern face recognition framework \cite{taigman2014deepface,schroff2015facenet,liu2017sphereface,Deng_2019_CVPR}.
Significant differences in the 1v1 results are observed by using templates with different degrees of external features involved, as illustrated in Fig. \ref{fig:intro_1}.
An open problem arises: \textit{is there an optimal template such that the produced face region gives the best recognition performance?}
Specifically, it remains unknown whether fewer backgrounds or irrelevant textures to face (e.g., hair, forehead) benefit face recognition. Besides, it is unclear whether the optimal template generalizes well across various conditions including the pose, age and illumination.


{}

Instead of manually designing  templates, we propose to automate the process of finding the optimal template for recognition.
To this end, we decompose differences of templates into  \textit{vertical shift} and \textit{crop size}, and construct a well-defined discrete searching space. We call the \textit{vertical shift} and \textit{crop size} pair an alignment policy.
The equivalence relation of the alignment policy and the template is described and proved in Section Face Alignment, and illustrated in Fig. \ref{fig:align}.
The template searching space is thus projected to the cropping box space spanned by \textit{vertical shift} and \textit{crop size}.

A straightforward way to search for the template is using the grid search.
However, grid search is inefficient and costly.
For example, the total size of searching space in our work is 93 and the grid search for the optimal template on the dataset like CASIA \cite{yi2014learning} is rather time-consuming (costs about \textbf{9102} GPU hours with 8 Tesla V100 GPUs).

In this paper, we propose an evolution-based method named Face Alignment Policy Search (FAPS) to efficiently searches for the optimal template.
FAPS jointly trains a population of models with evolving templates. Inspired by PBT \cite{jaderberg2017population}, we reuse the partially trained weights to accelerate
the searching procedure, as training from scratch on a large-scale dataset is time-consuming.
To improve the generality of the partially trained model, we set the upper bound of search space as \textit{SuperROI} such that the models have the knowledge of all the facial parts and can concentrate on the more informational area.
The original \textit{explore} in PBT mainly considers perturbing the hyperparameter from a better-performing population or resampling new hyperparameter from originally defined distribution, while ignores the relations among different templates in our problem.
To accelerate the discovering of better \textit{crop size} and \textit{vertical shift}, we propose \textit{Intersection based Crossover} to combine the strength of well-performing templates (Fig.  \ref{fig:crossover}).

Until now, searching for alignment in face recognition is less-studied and there exists no common protocol for evaluation, thus we introduce a well-designed benchmark(including LFW \cite{huang2008labeled}, AgeDB-30 \cite{moschoglou2017agedb} and MultiPIE \cite{gross2010multi}, \textit{etc.}) to evaluate the searched face crop template.

Our main contributions include: (i) To the best of our knowledge, we explore and highlight the effects of alignment templates on face recognition for the first time.
(ii) We construct a well-defined searching space by decomposing the template searching into \textit{crop size} and \textit{vertical shift} searching, and propose an efficient method named FAPS for template searching. (iii) A well-designed benchmark is proposed to evaluate the searched policy. Extensive experiments on the proposed benchmark validate the efficacy of FAPS.



\section{Background}
\noindent \textbf{Face Alignment}
is used to align faces to a unified distribution and reduce the geometric variations. The most commonly adopted way  is applying a 2D affine transformation to calibrate facial landmarks to predefined 2D \cite{Wang2018,Deng_2019_CVPR,wang2017normface,liu2017sphereface} or 3D templates  \cite{Taigman2014, guo2020towards}.

Besides the affine transformation, some other works learn non-rigid transformations. For example, ReST \cite{wu2017recursive} introduces a recursive spatial transformer to learn complex  transformation.
\cite{Zhou2018} use local homography transformations estimated by a rectification network to rectify faces. These methods aim for alignment-free through learning alignment jointly with the recognition network in an end-to-end fashion. Despite their achievements, additional computational cost and loss of identity information limit their usage in real-world applications.


Apart from the types of transformation, another critical element of  alignment is how to design a proper facial template.
Some early works \cite{lapedriza2005external,Andrews2010Internal} have observed performance improvements when including some external face features (\textit{i.e.}, hair, chin and face outline) compared to using internal face features alone (\textit{i.e.}, eyes, nose and mouth).
One optimal solution is to apply multi-patches methods \cite{Sun2014,Sun2014a,SunWT14a,LiuDBH15} which process an image via multiple templates and dump them to different recognition models.
Although this strategy improves  performances, it requires too much additional computational costs and carefully designed ensemble methods.
In our work, we compare the performance of a set of templates and aim to find the optimal one for the face recognition task.
\noindent \textbf{Hyperparameter Optimization}.
As face alignment policy is a hyperparameter for face recognition, our work closely correlates with the hyperparameter optimization\cite{feurer2019hyperparameter} problem which automatically tunes the hyperparameters.
An RL-based method called AutoAugment \cite{cubuk2019autoaugment} is proposed to train a controller to search for the best data augmentation policy based on specific datasets and models.
Apart from the RL-based methods, evolution-based methods \cite{jaderberg2017population,ho2019population} spring recently.
For example, PBT \cite{jaderberg2017population} jointly trains a population of models and searches for their hyperparameters with evolution to improve the models' performances.
\textit{Exploit} and \textit{explore} are the two most important strategies of PBT. \textit{Exploit} is responsible for copying better weights and hyperparameters from a well-performing model to the inferior one. \textit{Explore} creates new hyperparameters for the poor-performing model by either resampling new hyperparameters from the originally defined prior distribution or perturbing the copied hyperparameters from a well-performing model.
These two strategies make PBT faster and more effective.

In this work, inspired by PBT, we develop a novel evolution-based method named FAPS to search for a better face alignment strategy. The \textit{exploit}  and \textit{explore} from PBT are also adopted in our method.



\section{Methodology}
In this section, we first review the face alignment process via 2D affine transformations and demonstrate that template searching can be decomposed into searching \textit{crop size} and \textit{vertical shift}.
Then we detail the proposed FAPS.

\subsection{Face Alignment}\label{alignment_analysis}
\begin{figure}[htb!]
    \centering
    \includegraphics[width=0.48\textwidth]{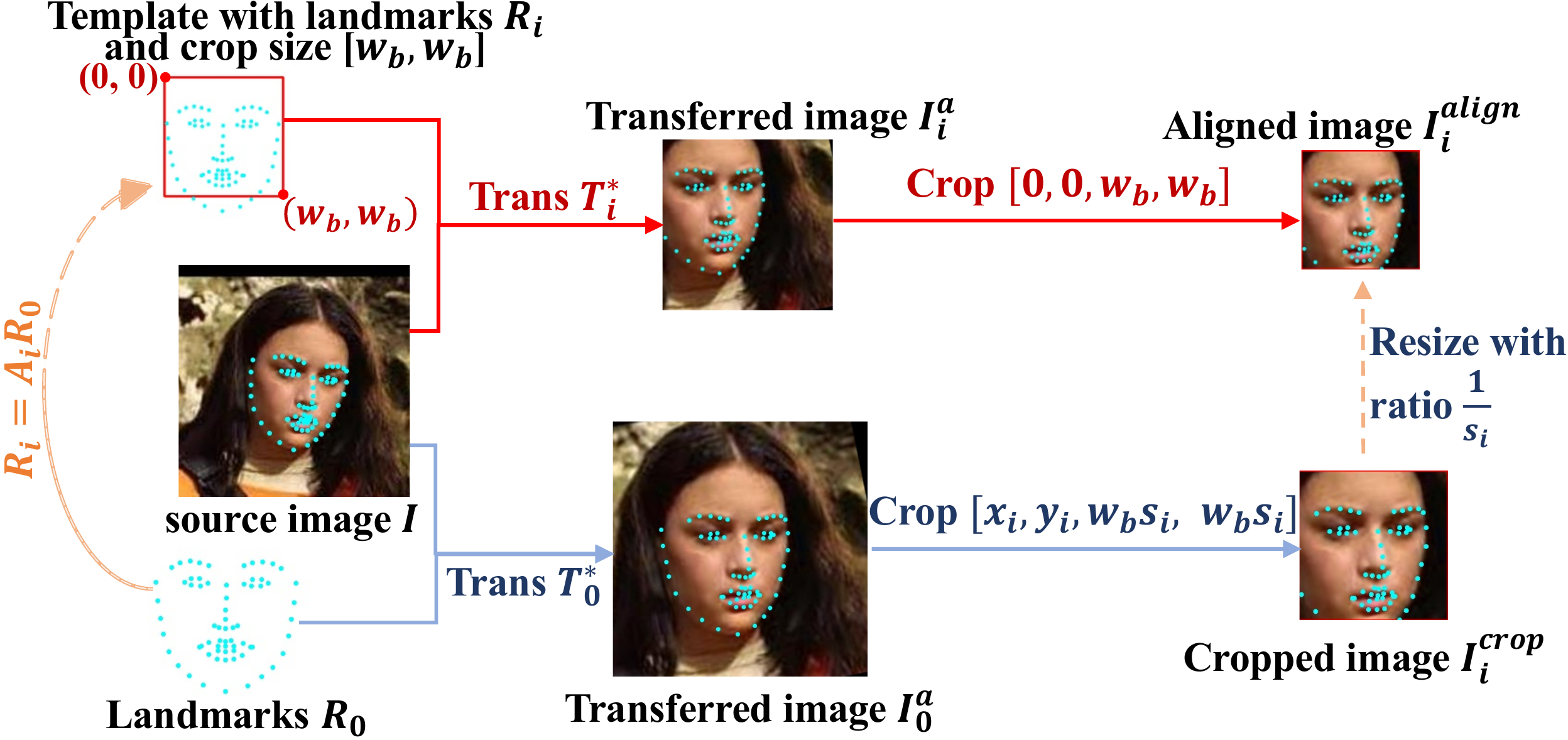}
    \caption{\normalsize{An overview of the face alignment process. Assuming we have a template with landmarks $R_i$ and cropping rectangle from point (0, 0) to point $(w_b, w_b)$. $R_i$ can be transferred from $R_o$ by  scaling $s_i$ and shifting $[x_i, y_i]$, \textit{i.e.,} $
      R_i =   \mathbf{A}_i R_0 =
    \left[
 \small{\begin{matrix}
  s_i & -s_i & x_i \\
    s_i& s_i & y_i \\
  0 & 0 & 1
  \end{matrix}}
  \right] R_0
$.
The source image $\mathbf{I}$ is transferred to $\mathbf{I}_0^a$ and $\mathbf{I}_i^a$ based on landmarks $R_0$ and $R_i$ respectively. We prove that the result $\mathbf{I}_i^{align}$ aligned by the current template is the same as resizing cropped image $\mathbf{I}_0^{crop}$. Therefore, the aligned image from an arbitrary template can be got by cropping and resizing from the same image $\mathbf{I}_0^a$.}}
    \label{fig:align}
\end{figure}

We define one alignment template as  a composition of landmarks $R_i$ with cropped area $[0, 0, w_b, w_b]$ (a $w_b\times w_b$ rectangle with top left point [0, 0]). In this work, facial landmarks in all templates share the same shape. To be more specific, any $R_i$ can be transformed from one base landmarks $R_0$  by scaling $s_i$ and shifting $x_i, y_i$ over the x, y axis respectively as shown in Fig. \ref{fig:align}.

One face image $\mathbf I$ is aligned to landmarks $R_i$ by a 2D affine transformation $\mathbf T$.
 Denote $\mathbf{I}^a_i$ as the transferred image based on landmarks $R_i$.  We seek an optimized affine transformation matrix $\mathbf{T}_i^*$ to transfer a face image $\mathbf I$ to $\mathbf{I}^a_i$. It can be proved that $\mathbf{T}_i^*=\mathbf{A}_i\mathbf{T}_0^*$.
Then we have $\mathbf{I}^a_i = \mathbf{T}^*[\mathbf I] = \mathbf{A}_i\mathbf{T}_0^*[\mathbf I] = \mathbf{A}_i\mathbf{I}^a_0$, which shows that the transferred image based on landmarks $R_i$ can be achieved by performing transformation $\mathbf{A}_i$ on the $\mathbf{I}^a_0$. The final aligned image is the area $[0, 0, w_b, w_b]$ of transferred image $\mathbf{I}^a_i$, which is given by the following steps: 1) Transfer image $\mathbf I$ to $\mathbf{I}^a_0$ based on the base landmarks $R_0$. 2) Crop the image with area $[x_i, y_i, w_b\cdot s_i, w_b\cdot s_i]$.
3) Resize the area by size $[w_b, w_b]$.

Therefore, instead of designing various templates and aligning a face multiple times,  we simplify the processes by aligning once by the base template $R_0$ and operating (crop + resize) on the same image $\mathbf{I}^a_0$.  In our implementation, landmarks in all templates are placed to be horizontally symmetric, which makes $x_i=0$. Let $m_i=w_b\cdot s_i$, $\delta_i = y_i/s_i$, our target now is to find the  optimal $m^*, \delta^*$.
We call $\textbf{\textit{p}} = \{m, \delta\}$ an alignment policy and each  policy  represents a corresponding  template.


\subsection{Search Space}\label{sec:search_space}

\begin{figure}
  \centering
  {\includegraphics[width=0.35\textwidth]{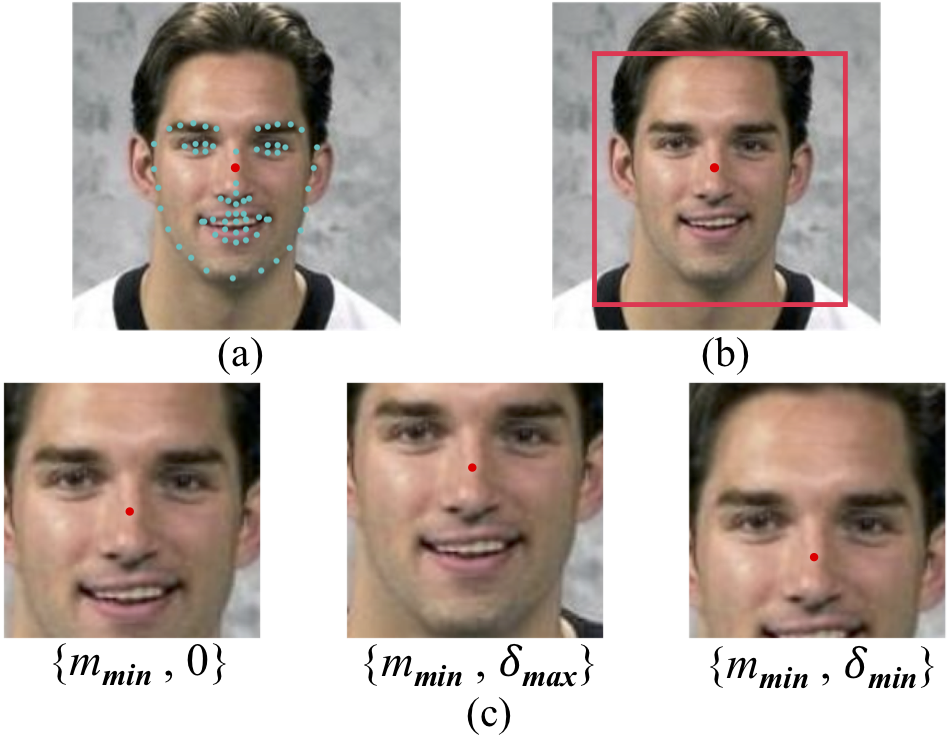}}

  \caption{\normalsize{An overview of the search space: (a) The face image is $300\times300$ after aligned with base landmarks $R_0$. The red landmark point is placed in the center of the canvas. (b) The red box ($\{m=m_{max}, \delta=0\}$) shows the upper bound of the search space.  (c) The input facial image varies rapidly with different $\delta$ and fixed crop size $m_{min}$. When  $\delta$ is 0, indispensable facial parts (eyes, nose, mouth) and half of the forehead and chin are kept. The forehead is almost removed when  $\delta=\delta_{max}$. When setting $\delta$ to $\delta_{min}$, the forehead is well-preserved while the chin is dropped.}
  }
    \label{fig:search_space}
\end{figure}

To facilitate the search process, we place the base face landmarks $R_0$ to a $300\times 300$ canvas with the mid-point of the nose (red point in Figure \ref{fig:search_space}(a)) at the center. We denote this template as $T_p$. After aligning an image to $R_0$, FAPS searches for the optimal region to simulate the effects of applying different templates. A candidate region is determined by 1) \textit{crop size} $m$ which controls the tightness of cropped face and 2) \textit{vertical shift} $\delta$  which controls the center of cropped area. Some examples are presented in Fig. \ref{fig:search_space}(c).

Denote $\mathcal{P}$ as the union of all candidate $\textbf{\textit{p}}$, \textit{i.e.,} the search space.  We define the search space as follows: With upper bound $m_{max}$ and $\delta=0$,  the selected region is able to cover both internal and external face features (Fig. \ref{fig:search_space}(b)). While with $m_{min}$ and $\delta=0$, only indispensable facial parts (eyes, nose, mouth) are kept as shown in Fig. \ref{fig:search_space}(c).

Through the variation of vertical shift $\delta$, some facial features are dropped  and some new features are included in the input. When $m$ is set to the smallest scale $m_{min}$, this phenomenon becomes more obvious (Fig. \ref{fig:search_space}(c)). If $\delta$ is set to the maximum value $\delta_{max}$, only the eyebrows are preserved, the forehead is almost omitted. When $\delta$ is set to the minimum value $\delta_{min}$, only the mouth is preserved, the chin is dropped. With such an
extreme setting of $\delta$, the importance of different facial areas can be discovered. \\



\subsection{Search Strategy}\label{sec:search_strategy}



Denote the recognition model as $f$ and its weights as $w$, we represent model trained with images aligned by $\textbf{\textit{p}}$ as $f(w|\textbf{\textit{p}})$. Let $\mathcal{L}_{train}$ and $\text{ACC}_{val}$ be the training loss and validation accuracy, respectively. The process of finding the optimal alignment policy can be formulated as:

\begin{align}
        \textbf{\textit{p}}^* & = \text{argmax}_{\textbf{\textit{p}}\in \mathcal{P}}   \text{ACC}_{val}(f(w^*|\textbf{\textit{p}})) \label{eq:simple_format1} \\
     \textit{s.t.} & \  w^* = \text{argmin}_w\ \mathcal{L}_{train}\ {f(w|\textbf{\textit{p}})}  \label{eq:simple_format2}
\end{align}

\begin{figure*}[htb!]
    \centering
    \includegraphics[width=0.88\textwidth, height=0.30\textwidth]{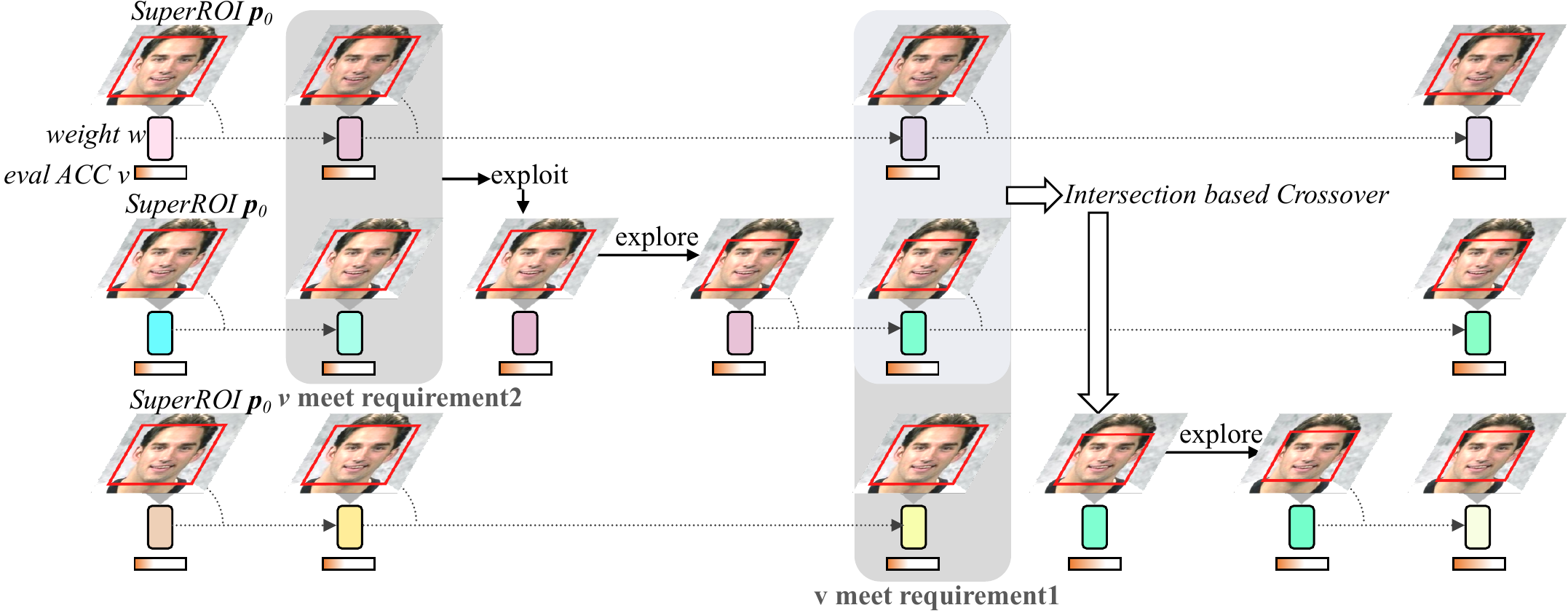}
    \caption{\normalsize{Overview of the proposed FAPS. {\normalfont We first initialize a fixed population of models with \textit{SuperROI} $\textbf{\textit{p}}_0$. After each epoch, each model's accuracy $v$ on the validation set is calculated.
    If an under-performing model meets \textit{requirement1}, the \textit{Intersection based Crossover} will be operated on the model. Then a new alignment policy is generated by combining the policies of two well-performing models.
    If an inferior model meets \textit{requirement2},  \textit{exploit} and \textit{explore} will be performed. To be more specific, model weights are copied by those of a superior model and new alignment policy is generated by disturbing a superior policy.}}}
    \label{fig:main_method}

\end{figure*}

To find the optimal solution, the trivial approach like grid search is to traverse all possible $\textbf{\textit{p}}$. In this way, model $f$ needs to be trained $|\mathcal{P}|$ times, which is time-consuming and inefficient.
Inspired by Population based Training (PBT)\cite {jaderberg2017population}, we train a fixed population of models with different $\textbf{\textit{p}}$ in parallel. The ``exploit-and-explore" procedure is applied to the worse performing models at a certain interval, where the inferior model clones the weight of better performing model and updates the alignment policy through perturbing this well-performing model's $\textbf{\textit{p}}$. The model can be trained with a new $\textbf{\textit{p}}$ without reinitialized. The total
computation is largely reduced to a single optimization process (Fig. \ref{fig:main_method}).

\subsubsection{SuperROI}
To improve the generality of partial trained model when cloning the weights,  we initialize $\textbf{\textit{p}}$ to $\{m_{max}, 0\}$ as shown in Fig. \ref{fig:search_space} (b), $\textit{i.e.}$, an initialized Region of Interest (ROI) containing all internal features (eyes, nose and mouth) and external features (jaw-line, ears, part of the hair, \textit{etc.}). Under this setting, beginning models can have the capacity to handle information from all facial parts. When switching to other policies, the facial region can be a part of the initial one and no new facial parts are introduced. Models only need to learn the trade-offs from current features, \textit{i.e.}, learn to focus on remaining facial parts and ignore removed ones.
This process shares the spirit of the supernet in Neural Architecture Search \cite{chen2019detnas,guo2019single,chu2019fairnas}, consequently, we name  $\textbf{\textit{p}}_0=\{m_{max}, 0\}$ as \textit{SuperROI}.

\subsubsection{Intersection based Crossover}\label{sec:crossover}
\begin{figure}[htb!]
    \centering
    \includegraphics[width=0.38\textwidth]{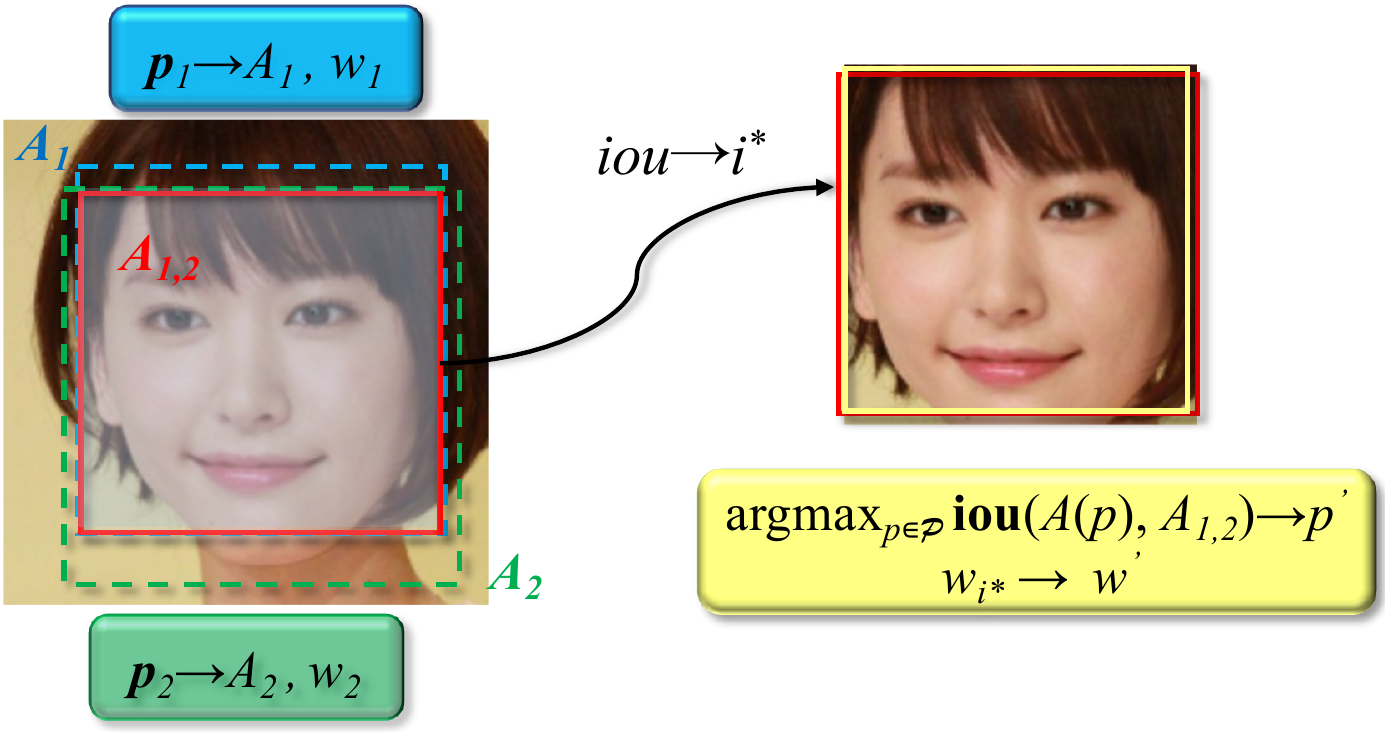}
    \caption{\normalsize{Illustration of \textit{Intersection based Crossover}. $\textbf{\textit{p}}_1 \text{ and } \textbf{\textit{p}}_2$ are alignment policies of two well-performing populations. Their corresponding regions are $A_1$ and $A_2$, $A_{1,2} = A_1 \cap A_2$ represents the shared area ((red rectangle)). Our \textit{Intersection based Crossover} finds a  policy $p^\prime$ which has the largest IOU scores with $A_{1,2}$ (yellow rectangle). As a result, $p^\prime$ inherits the intersection area.
    The $\textbf{iou}$ function decides whose weight can be cloned to the inferior model. The IOU score of $A_1 \text{ and } A^\prime$ is larger, hence $w_1$ is chosen.}}
    \label{fig:crossover}
\end{figure}

The original \textit{explore} of PBT either re-samples new hyperparameter directly from the originally defined prior distribution or perturbs the current hyperparameter from a well-behaved population to upgrade the weak-behaved population. The former strategy, which resembles random search \cite{bergstra2012random}, can relieve the problem of local minima but cannot guarantee qualities of sampled hyperparameters. The later strategy is analogous to the mutation in genetic algorithms and has a high probability of finding better hyperparameter. However, it generates new hyperparameter depending on one particular hyperparameter each time instead of
hyperparameters of well-behaved populations, which may lead to unstable results.
Besides the above hyperparameter generation methods, the common trend of well-behaved ones is  not fully utilized.
Inspired by crossover in genetic algorithms \cite{spears1993crossover}, we propose \textit{Intersection based Crossover} to facilitate the discovering of better alignment policy $\textbf{\textit{p}}$ during search (Fig. \ref{fig:crossover}).
Suppose there exist two well-performing policies $\textbf{\textit{p}}_1=\{m_1, \delta_1\},  \textbf{\textit{p}}_2=\{m_2, \delta_2\}$ and the corresponding facial areas are $A_1, A_2$ respectively. Their intersection area $A_{1,2} = A_1 \cap A_2 $ is highly possible to contain rich facial information that benefits face recognition.  Policies generated by trivial crossover ($\{m_1, \delta_2\}$ and $\{m_2, \delta_1\}$) can possibly represent regions that differ a lot from both $A_1, A_2$, which therefore fail to cover the intersection area.
Instead, \textit{Intersection based Crossover}
finds the policy whose region has the largest similarity with $A_{1,2}$. Denote $A(\textbf{\textit{p}})$ as the face region represented by policy $\textbf{\textit{p}}$  and $\textbf{iou}(A(\textbf{\textit{p}}), A_{1,2}) = \frac{A(\textbf{\textit{p}}) \cap A_{1,2}}{A(\textbf{\textit{p}}) \cup A_{1,2}}$,  we update the policy $\textbf{\textit{p}}$ and model weights $w$ by Eq.\ref{eq:iou_3} and Eq.\ref{eq:iou_4}:

\begin{equation}
	\textbf{\textit{p}}^\prime  \leftarrow \text{argmax}_{\textbf{\textit{p}}\in \mathcal{P}} \textbf{iou}(A(\textbf{\textit{p}}), A_{1,2})
\label{eq:iou_3}
\end{equation}
\begin{equation}
	w^\prime \leftarrow w_{i^*},\  s.t.\   i^* = \text{argmax}_{i  \in \{1,2\}}\ \textbf{iou}(A(\textbf{\textit{p}}^\prime), A_i)
\label{eq:iou_4}
\end{equation}


\subsection{Implementation} \label{sec:implementation}

The alignment template search process is elaborated in Algorithm \ref{alg:A}.  The details of the main function are below:\\
\textbf{Step:} In each step, we train the model in one epoch through SGD with ArcFace loss \cite{Deng_2019_CVPR}.\\
\textbf{Eval:} We evaluate the current model on our validation set, the verification rate is calculated as the validation accuracy.\\
\textbf{Ready:} A model is ready to go through the exploit-and-explore or \textit{Intersection based Crossover} process once 1 epoch has elapsed.\\
\textbf{Requirement1:} The model's validation accuracy $v$ is between the bottom $1/4$ and $3/8$ of the population.\\
\textbf{Requirement2:} The model's validation accuracy $v$ is in the bottom $1/4$ of the population.\\
\textbf{Exploit:} Get the weight $w$ and alignment policy $\textbf{\textit{p}}$ of a model that has validation accuracy $v$ in the top $1/4$.\\
\textbf{Explore:} See Algorithm \ref{alg:explore} for the \textit{explore} function. For $m$ and $\delta$, we either perturb the original value or uniformly resample them from all possible values.\\
\textbf{Intersection based Crossover:} We choose two well-performing models $f(w_1|\textbf{\textit{p}}_1)$ and $f(w_2|\textbf{\textit{p}}_2)$ whose validation accuracies are in the top $1/4$ to generate the new alignment policy $\textbf{\textit{p}}^\prime$. If $\textbf{\textit{p}}^\prime$ is already deployed by the current models, an extra \textit{explore} will be applied to $\textbf{\textit{p}}^\prime$.

\begin{algorithm}[]
\small{
\caption{\small{Face Alignment Policy Search(FAPS).}}
\label{alg:A}
\begin{algorithmic}[1]
\Require{Current policy search space $\mathcal{P}$, \textit{SuperROI} $\textbf{\textit{p}}_0 = \{m_{max}, 0\}$, population size of models $\textit{N}$.}
\State \!\!\!\! Initialize $\textit{N}$ models $f(w|\textbf{\textit{p}}_0)$ \\
    \textbf{for} {each model $f(w|\textbf{\textit{p}}_0)$ (asynchronously in parallel)} \\
      \quad \!\!\! \textbf{while} {not end of training} \\
      \quad \!\!\! \quad \!\!\! $w\leftarrow \text{step}(w|\textbf{\textit{p}})\quad $ \!\!\!\!\!\!   \Comment{train current model with policy $\textbf{\textit{p}}$} \\
      \quad \!\!\! \quad \!\!\! $v\leftarrow {ACC}_{val}(f(w|\textbf{\textit{p}}))$  \Comment{evaluation} \\
      \quad \!\!\! \quad \!\!\! \textbf{if} {ready($f, v$)} \textbf{then}\\
      \quad \!\!\! \quad \!\!\! \quad \!\!\! check $v$'s performance among all models \\
      \quad \!\!\! \quad \!\!\! \quad \!\!\! \textbf{if} {$v$ meets \textit{requirement1} } \textbf{then} \\
      \quad \!\!\! \quad \!\!\! \quad \!\!\! \quad \!\!\! generate $w^\prime,  \textbf{\textit{p}}^\prime$ via  \textit{Intersection based Crossover} \\
      \quad \!\!\! \quad \!\!\! \quad \!\!\! \quad \!\!\! \textbf{If} {$\textbf{\textit{p}}^\prime$ doesn't exist currently}  \textbf{then} \\
      \quad \!\!\! \quad \!\!\! \quad \!\!\! \quad \!\!\! \quad \!\!\! $w, \textbf{\textit{p}} \leftarrow w^\prime, \textbf{\textit{p}}^\prime $  \\
      \quad \!\!\! \quad \!\!\! \quad \!\!\! \quad \!\!\! \textbf{else} \\
      \quad \!\!\! \quad \!\!\! \quad \!\!\! \quad \!\!\! \quad \!\!\! $w, \textbf{\textit{p}} \leftarrow \textit{explore}(w^\prime, \textbf{\textit{p}}^\prime)$ \\
      \quad \!\!\! \quad \!\!\! \quad \!\!\! \textbf{elif} {$v$ meets \textit{requirement2} } \textbf{then} \\
      \quad \!\!\! \quad \!\!\! \quad \!\!\! \quad \!\!\! get $w^\prime, \textbf{\textit{p}}^\prime$ through \textit{exploit} \\
      \quad \!\!\! \quad \!\!\! \quad \!\!\! \quad \!\!\! $w, \textbf{\textit{p}} \leftarrow \textit{explore}(w^\prime, \textbf{\textit{p}}^\prime) $ \\
      \quad \!\!\! \quad \!\!\! \quad \!\!\! update model populations with new $f(w|\textbf{\textit{p}})$ \\
   return \textbf{\textit{p}} with highest $v$ among training
\end{algorithmic}
}
\end{algorithm}

\begin{algorithm}
\small{
\caption{\small{The FAPS explore function. When revising the alignment policy based on the current one, the change value is amplified by magnitude parameters.}}
\label{alg:explore}
\begin{algorithmic}[1]
\Require {current alignment policy $\textbf{\textit{p}}=\{m, \delta\}$, \textit{SuperROI}, magnitude parameters $\textbf{\textit{s}} = \{s_{m}, s_{\delta}\}$} \\
    \textbf{for}{ \textit{param} in $\textbf{\textit{p}}$} \\
    \quad \!\!\! \textbf{if} {random(0, 1) \textless 0.2} \textbf{then}\\
    \quad \!\!\! \quad \!\!\! random sample \textit{param} uniformly from search space \\
    \quad \!\!\! \textbf{else} \\
    \quad \!\!\! \quad \!\!\! \textit{level} = [0,1,2,3] with probability [0,1, 0.3, 0.3, 0.3] \\
    \quad \!\!\! \quad \!\!\! \textbf{if} {random(0,1) \textless 0.5} \textbf{then}\\
    \quad \!\!\! \quad \!\!\! \quad \!\!\! $param = param - level \times  \textbf{\textit{s}}_{param}$ \\
    \quad \!\!\! \quad \!\!\! \textbf{else} \\
    \quad \!\!\! \quad \!\!\! \quad \!\!\! $param = param + level \times  \textbf{\textit{s}}_{param}$ \\
    \quad \!\!\! \quad \!\!\! Clip \textit{param} to stay within \textit{SuperROI}
\end{algorithmic}
}
\end{algorithm}


\section{Experiments}

\subsection{FAPS Benchmark} \label{sec:faps_benchmark}

To evaluate the influence of different alignment templates and the effectiveness of the proposed FAPS, we introduce a well-designed benchmark which includes searching set, training set, validation set and test set. We present our proposed benchmark in Table \ref{table:benchmark}.

The scale of the training dataset is an important factor for face recognition. We separately employ CASIA \cite{yi2014learning} and MS-Celeb-1M \cite{guo2016ms} as middle-scale and large-scale training and searching datasets.
For CASIA, we use the full dataset as the searching data and training data.
For MS-Celeb-1M,
we use MS-Celeb-1M-v1c \footnote{\url{http://trillionpairs.deepglint.com/overview}} which remains the completeness of facial images and is highly clean for training.
Searching on the MS-Celeb-1M-v1c directly requires too many computational resources.
To reduce the searching time, we sample 30000 identities with 30 images per identity from the whole dataset. This subset is named \textit{Reduced MS-Celeb-1M-v1c}.


Considering different data distributions and characteristics among datasets of the searching set, we enrich the variety of validation set to ensure the generalization of searched policies. The validation set is designed considering the main challenges of face recognition like age, pose and illumination variations. As a result, we build a validation dataset named Cross Challenge in the Wild (CCW), the images are from three datasets in unconstrained environments: LFW\cite{huang2008labeled}, AgeDB-30\cite{moschoglou2017agedb} and  CPLFW \cite{zheng2018cross}.


The test set including LFW, AgeDB-30, CALFW \cite{zheng2017cross}, CPLFW, MultiPIE \cite{gross2010multi} and IJB-A  \cite{klare2015pushing}.
More details of the benchmark are presented in Appendix.







\subsection{Experimental Settings}

We detect the faces by adopting the s3fd detector \cite{zhang2017s3fd}
and localize 68 landmarks via FAN \cite{bulat2017far}.  Images are  affined according to the predefined $300\times300$ average face template $T_p$ as shown in Fig. \ref{fig:search_space}(a). Faces are cropped and resized with different alignment policies for searching, but with consistent policies for training, validation and testing. The cropped faces are then resized to $112\times112$.


The widely used ResNets \cite{he2016deep} with 
embedding structure \cite{Deng_2019_CVPR} are employed as our recognition networks.
The embedding dimension is set to 512. To accelerate the searching process, ResNet18 is adopted as the searching network. ResNet50 is used to train on the training set. ArcFace \cite{Deng_2019_CVPR}  is served as
the loss function during searching and training. We implement FAPS with PyTorch  \cite{paszke2019pytorch} and Ray Tune \cite{moritz2018ray}.

During searching, the population size of models $\textit{N}$ is set to 8.
The crop size $m_{max} \text{ and } m_{min}$ are set to 232 and 160, respectively. The vertical shift $\delta_{max} \text{ and } \delta_{min}$ are 24 and -32. We set the magnitude parameter of crop size $s_{m}=8$ and the magnitude parameter of vertical shift $s_\delta=4$. Under this setting, we have 93 candidates in
the template searching space $\mathcal{P}$. More setting details are shown in Appendix.

\begin{table}
\begin{center}
\small
\begin{tabular}{c|c|c}
\hline
Benchmark & CASIA & MS-Celeb-1M-v1c \\
\hline
Searching Set & CASIA & \textit{Reduced MS-Celeb-1M-v1c} \\
\hline
Training Set  &  CASIA & MS-Celeb-1M-v1c \\
\hline
\multirow{1}{*}{Validation Set}  & CCW  & CCW  \\
\hline
\multirow{5}{*}{Test Set}  & LFW & LFW \\
          & AgeDB-30 & AgeDB-30 \\
          & CPLFW  & CPLFW \\
          & CALFW & CALFW \\
          & MultiPIE & MultiPIE \\
          &  & IJB-A \\
\hline
\end{tabular}
\caption{\normalsize{FAPS Benchmark}}
\label{table:benchmark}
\end{center}

\end{table}

\subsection{Compared Methods}

For comparison, we map the widely-used 5-points template presented in ArcFace  \cite{Deng_2019_CVPR} to the predefined $300\times 300$ template $T_p$, which results in  policy $\textbf{\textit{p}}=\{190,-7\}$. Another 25-points alignment template utilized by MFR \cite{guo2020learning} and works \cite{zhu2019large,guo2018face} is mapped to $\{198, -15\}$. We call policy $\{m_{min}, 0\} = \{160, 0\}$ the \textit{TightROI} which involves few external face features. \textit{SuperROI} as well as the aforementioned three policies are treated as compared policies. We further compare the proposed FAPS with the spatial-transform based methods ReST \cite{wu2017recursive} and GridFace\cite{Zhou2018}.
Fig. \ref{fig:experiment_analysis} shows some aligned faces with different policies.
ReST and GridFace coupled alignment with recognition network, they can hardly be mapped into our search space.


\subsection{Searching on CASIA}\label{sec:casia_}

In this section, CASIA is used as the searching and training sets. The corresponding validation/test sets are presented in Table \ref{table:benchmark}. FAPS's searching process takes \textbf{131} GPU hours with 8 Tesla V100 GPUs. As a comparison, the grid search method with ResNet18 takes about \textbf{9102} GPU hours. 
With the searched alignment policy, we train the ResNet50 from scratch for 32 epochs. The learning rate is initialized by 0.1 and divided by 10 at epoch 20 and 28.

\begin{figure}[htb!]
    \centering
    \includegraphics[width=0.44\textwidth]{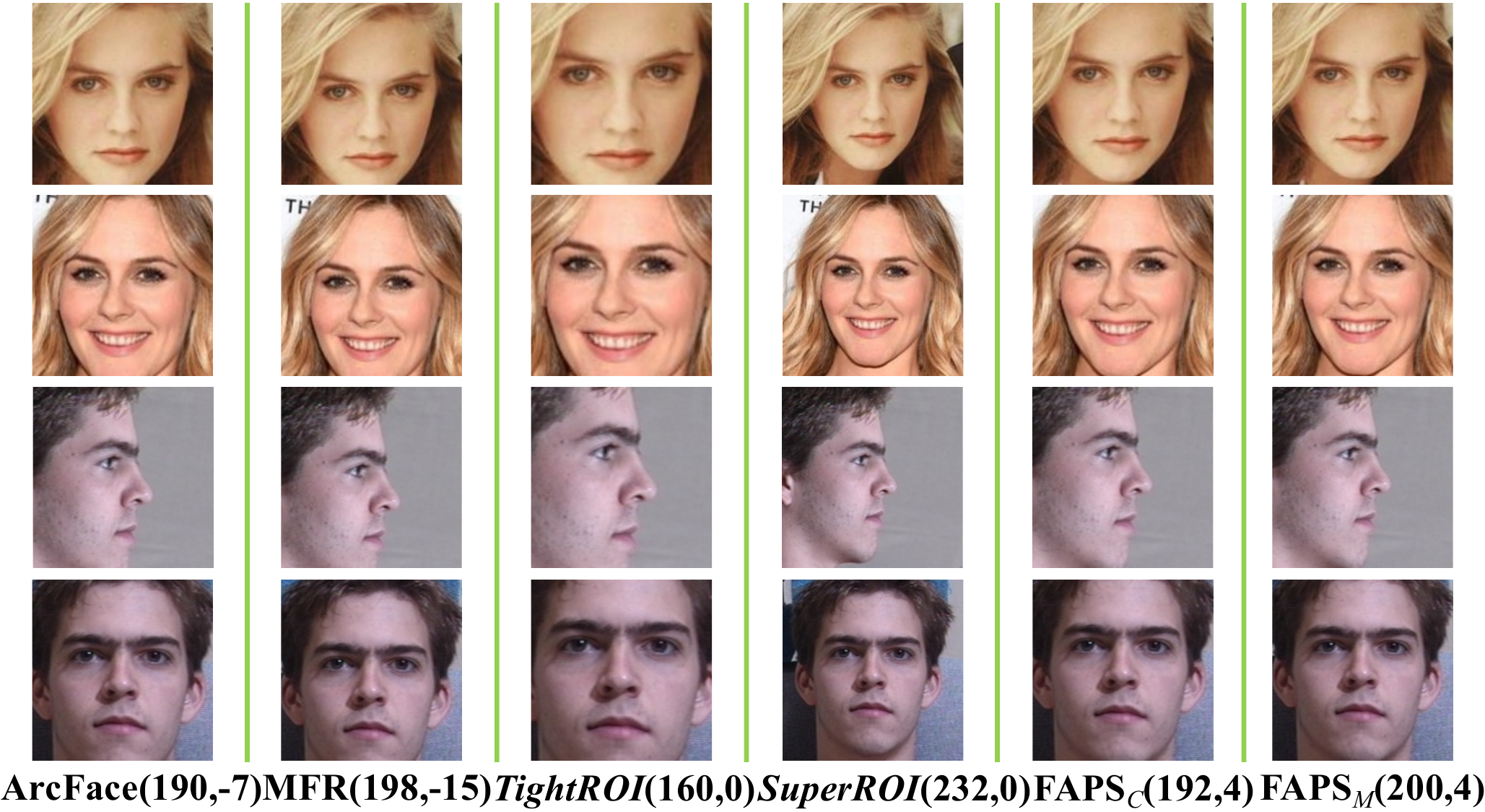}

    \caption{\normalsize{Face images aligned with different templates. {\normalfont The first two rows show faces of the same person in CALFW. Faces of the last two rows are from MultiPIE $\pm 90^{\circ}$ subset and MultiPIE $0^{\circ}$ subset respectively, they are the same identity as well.}
    }}
    \label{fig:experiment_analysis}

\end{figure}

Results are summarized in Table \ref{table:casia_r50_1} and Table \ref{table:ms1m_res2} (results of the baseline will be discussed in Ablation Study).
We denote the searched alignment policy FAPS$_C$(192,4).
Obviously, FAPS$_C$(192,4) surpasses the compared policies on all test datasets.
For example, on LFW, FAPS$_C$(192,4) outperforms all other policies, especially the \textit{TigthROI}. With the same training dataset, FAPS$_C$(192,4) achieves a 0.45\% improvement above ReST. On AgeDB-30 and CALFW, FAPS$_C$(192,4) shows significant improvements over the best results from compared policies by 0.78\% and 1.15\%.
As shown in Fig. \ref{fig:experiment_analysis}, FAPS$_C$(192,4) drops more hair than ArcFace's and MFR's but remains more chin. This indicates that hair is not helpful for face recognition with age challenge as people's hairstyles usually change during their lifetime, while the chin and the outline of chin remain unchanged.

For profile faces, FAPS$_C$(192,4) gains improvement over other compared policies on CPLFW, MultiPIE $\pm 75^{\circ}$ and MultiPIE $\pm 90^{\circ}$.
On the less challenging MultiPIE $\pm 60^{\circ}$, FAPS$_C$(192,4) performs as good as MFR and \textit{TightROI}.

These results show FAPS's searched alignment policy gains superiority over handcrafted ones for faces with large pose variations.
This mainly because profile faces are aligned to one side of the images (as shown in Fig. \ref{fig:experiment_analysis}). Policies with  too small crop sizes (\textit{e.g.}, \textit{TightROI}) filter out useful face features, while large crop sizes (\textit{e.g.}, \textit{SuperROI}) can bring irrelevant features and background noise. In contrast, our FAPS can find a trade-off and therefore can focus on key features.


\begin{table}
\begin{center}
\small
\resizebox{\columnwidth}{!}{
\begin{tabular}{c|c|c|c|c|c}
\hline
Training Set & Method & LFW &  AgeDB-30 & CALFW & CPLFW  \\
\hline
\multirow{6}{*}{CASIA} & ReST & 99.03 & - & - & - \\
\cline{2-6}
& ArcFace (190,-7) & 99.43 & 94.42 & 90.92 & 85.15\\
& MFR (198,-15)  & 99.43 & 94.47 & 91.15 & 84.75 \\
\cline{2-6}
& \textit{TigthROI} (160,0)   & 99.17 & 94.23 & 91.15 & 85.07 \\
& \textit{SuperROI} (232,0)  & 99.43 & 94.47 & 90.48 & 83.97 \\
\cline{2-6}
& baseline (184,4) & 99.45 & 95.03  & 91.07 &  \textbf{85.88} \\
& FAPS$_{C}$ (192,4) &  \textbf{99.48} &  \textbf{95.25} & \textbf{92.07} & 85.43 \\
\hline
\hline
\multirow{6}{*}{MS1M} & GridFace & 99.70 & - & - & - \\
\cline{2-6}
& ArcFace (190,-7)  & 99.72 & 98.02 & 95.23 & 87.98  \\
& MFR (198,-15)  & 99.77 & 97.78 & 95.47 & 87.28 \\
\cline{2-6}
& \textit{TigthROI} (160,0)  &  99.73 & 97.95 & 95.47 & 88.13 \\
& \textit{SuperROI} (232,0) & 99.77 & \textbf{98.25} & 95.47 & 88.05 \\
\cline{2-6}
& FAPS$_{C}$ (192,4) &  99.78 & 98.10 & \textbf{95.78} & 88.12 \\
& FAPS$_{M}$ (200,4) & \textbf{99.82} & 98.08 & 95.65 & \textbf{88.95} \\

\hline
\end{tabular}
}
\caption{\normalsize{Verification performance (\%) at different alignment policies with ResNet50 backbone.
MS1M: MS-Celeb-1M-v1c. FAPS$_{C}$(192,4) and FAPS$_{M}$(200,4) denote the policies searched on CASIA and \emph{Reduced MS-Celeb-1M-v1c}, respectively.
}}
\label{table:casia_r50_1}
\end{center}
\end{table}

\subsection{Searching on MS-Celeb-1M-v1c}
\begin{table}
\begin{center}
\small
\begin{tabular}{c|c|c|c|c}
\hline
Training Set & Method & $\  \pm 90^{\circ}$ \  & $\  \pm 75^{\circ}$   & $\  \pm 60^{\circ}$ \\
\hline
\multirow{6}{*}{CASIA} & ArcFace (190,-7)  & 89.5 & 97.0 & 99.3 \\
& MFR (198,-15)  & 91.2 & 97.7 & \textbf{99.7} \\
\cline{2-5}
& \textit{TigthROI} (160,0)  &  90.8  & 97.6 & \textbf{99.7} \\
& \textit{SuperROI} (232,0) & 90.7 & 97.1 & 99.3 \\
\cline{2-5}
& baseline (184,4) & 90.4 & 97.5 & 99.6	\\
& FAPS$_C$ (192,4) & \textbf{91.7} & \textbf{98.3} & \textbf{99.7}  \\
\hline
\hline
\multirow{6}{*}{MS1M} & GridFace & 75.4 & 94.7 & 99.2  \\
\cline{2-5}
& ArcFace (190,-7)  & 70.4 & 98.8  & \textbf{100.0}  \\
& MFR (198,-15)  & 71.9 & 98.9 & \textbf{100.0} \\
\cline{2-5}
& \textit{TigthROI} (160,0) & 68.7 &  98.4 & \textbf{100.0} \\
& \textit{SuperROI} (232,0) & 70.7 & 98.0 & 99.9 \\
\cline{2-5}
& FAPS$_C$ (192,4) & 74.6 & \textbf{99.0} & \textbf{100.0} \\
& FAPS$_M$ (200,4) & \textbf{76.6} & 98.8 & \textbf{100.0} \\
\hline
\end{tabular}
\caption{\normalsize{Rank-1 recognition rates (\%) for different poses at different alignment policies on MultiPIE with ResNet50 backbone. FAPS$_{C}$(192,4) and FAPS$_{M}$(200,4) denote the policies searched on CASIA and \emph{Reduced MS-Celeb-1M-v1c}, respectively.}}
\label{table:ms1m_res2}
\end{center}
\end{table}

In this section, \textit{reduced MS-Celeb-1M-v1c} is used as the searching data. 
The searching process takes \textbf{234} GPU hours with 8 Tesla V100 GPUs, while the grid search method   takes more than \textbf{4812} GPU hours. After the searching phase, we train ResNet50 for 16 epochs from scratch with the searched alignment policy on the full data,
with learning rate initialized as 0.1 and dropped by 10 at the 8th and 14th epochs.

Results are showed in Table \ref{table:casia_r50_1}, \ref{table:ms1m_res2}, \ref{table:ms1m_ijba}. When compared with other handcrafted alignment policies, FAPS's searched policy on \textit{Reduced MS-Celeb-1M-v1c} (FAPS$_M$(200,4)) outperforms other policies on almost all datasets.
On LFW, FAPS$_M$(200,4) outperforms the human-designed policies by at least 0.05\%. As the performance on LFW is almost saturated around 99.80\%, the improvement is nonnegligible. On CALFW, FAPS$_M$(200,4) outperforms other handcrafted alignment policies by almost 0.2\%. For profile faces, the searched policy FAPS$_M$(200,4) can obviously boost the performance on both CPLFW and MultiPIE $\pm 90^{\circ}$ by 0.82\% and 4.7\%. On the challenging dataset IJB-A, FAPS$_M$(200,4) achieves best verification and identification performance.

With the same training data MS1M,
our FAPS$_M$(200,4) achieves a 0.12\% improvement above GridFace on LFW.
On MultiPIE $\pm 90^{\circ}$, $\pm 75^{\circ}$ and $\pm 60^{\circ}$, FAPS$_M$(200,4) surpasses GridFace by clear margins. On IJB-A, FAPS$_M$(200,4) gains obvious improvement on verification accuracy(3.0\% and 7.3\%)
it also shows superiority on identification accuracy.

To further verify the generalization of our searched template, we train ResNet50 on MS-Celeb-1M-v1c with the policy FAPS$_C$(192,4) which searched on CASIA. When compared with handcrafted alignment policies, FAPS$_C$(192,4) also gains better performance on almost all the datasets while a little bit inferior to FAPS$_M$(200,4)'s. It shows improvements on LFW, CALFW, MultiPIE $\pm 90^{\circ}$, $\pm 75^{\circ}$ and gains comparable performance on CPLFW and MultiPIE $\pm 60^{\circ}$. On IJB-A, FAPS$_C$(192,4) boosts the verification accuracy with FAR at 0.001 and the Rank-1 accuracy. These results show the generalization of the searched alignment policies of FAPS. Once searched on one dataset, the searched policy can further improve the recognition performance when trained on different datasets.

On both CASIA and MS-Celeb-1M-v1c, the searched alignment policies gain better performance. It shows that compared to current human-designed alignment templates, the optimal one can be searched by FAPS to facilitate the face recognition performance. The searched alignment policy can also generalize across different training datasets. Moreover, although the searched alignment policy of MS-Celeb-1M-v1c is different from CASIA's, the input facial area decided by the two searched policies are almost overlapped (IOU 0.92). Almost all chin and part of the forehead are contained for both policies. The results show that adding proper external facial features is beneficial to recognition.

\begin{table}
\begin{center}
\small
\resizebox{\columnwidth}{!}{
\begin{tabular}{c|c|c|c|c}
\hline
Method $\downarrow$ & \multicolumn{2}{|c}{Verification} & \multicolumn{2}{|c}{Identification} \\
Metric $\rightarrow$ & @FAR = 0.01 & @FAR = 0.001 & @Rank-1 & @Rank-5 \\
\hline
GridFace & 92.1 $\pm$ 0.8 & 83.9 $\pm$ 1.4 & 92.9 $\pm$ 1.0 &  96.2 $\pm$ 0.5\\
\hline
ArcFace (190,-7)  & 94.5 $\pm$ 0.6 & 87.1 $\pm$ 1.4 & 93.1 $\pm$ 0.8 & 95.5 $\pm$ 0.4 \\
MFR (198,-15)  & 94.7 $\pm$ 0.6 & 88.6 $\pm$ 1.0 & 93.7 $\pm$ 0.7 & 96.0 $\pm$ 0.6 \\
\hline
\textit{TigthROI} (160,0) & 93.6 $\pm$ 0.8 & 82.1 $\pm$ 2.8 & 92.4 $\pm$ 0.7 &  95.0 $\pm$ 0.6  \\
\textit{SuperROI} (232,0) & 95.1 $\pm$ 0.7 & 87.4 $\pm$ 1.9 & 93.7 $\pm$ 0.8 & 95.8 $\pm$ 0.5 \\
\hline
FAPS$_C$ (192,4) & 94.8 $\pm$ 0.6 & 89.7 $\pm$ 1.4 & 93.8 $\pm$ 0.8 & 95.9 $\pm$ 0.5 \\
FAPS$_M$ (200,4) & \textbf{95.1 $\pm$ 0.6} & \textbf{91.2 $\pm$ 0.6} & \textbf{94.1 $\pm$ 0.7} & \textbf{96.4 $\pm$ 0.4} \\
\hline
\end{tabular}
}
\caption{\normalsize{Results on IJB-A with searched policies FAPS$_C$(192,4) and FAPS$_M$(200,4). The training set is MS-Celeb-1M-v1c.}}
\label{table:ms1m_ijba}
\end{center}
\end{table}

\subsection{Ablation Study}

\subsubsection{Effectiveness of Intersection based Crossover}
\label{sec:ablation_IOU}
We first evaluate \textit{Intersection based Crossover}, the method we proposed to facilitate the discovering of better alignment policies. To analyze its impact, we search for the CASIA's alignment policy under the same setting as that in section Searching on CASIA, but without \textit{Intersection based Crossover}. The searched policy without \textit{Intersection based Crossover} is named baseline. The results are summarized in Table \ref{table:casia_r50_1} and \ref{table:ms1m_res2}. The policy FAPS$_C$(192,4) discovered with \textit{Intersection based Crossover} shows better results compared to the baseline at almost all test datasets. Specifically, FAPS$_C$(192,4) outperforms baseline by 1.0\% at CALFW, 1.3\% and 0.8\% at MultiPIE $\pm 90^{\circ}$ and $\pm 75^{\circ}$. At CPLFW, FAPS$_C$(192,4) is slightly inferior to baseline. The reason may be CPLFW has more background noise and occlusion than MultiPIE. The facial area decided by FAPS$_C$(192,4) is a bit larger than baseline's, which means more noise involved.

\section{Conclusions}
In this paper, we explore the effects of different alignment templates on face recognition and propose a fast and effective alignment policy search method named FAPS. The searched templates via FAPS achieve better recognition performance compared to human-designed ones on multiple test datasets and generalize across different training datasets.
Besides, our searched templates reveal that except for the internal facial features like eyes, nose and mouth, external features like chin and jawline are helpful for face recognition. This also sheds some light on the further development of face recognition.

\section{Acknowledgments}
This work was supported in part by the National Key Research \& Development Program (No. 2020AAA0140002), Chinese National Natural Science Foundation Projects \#61876178, \#61806196, \#61976229.

\bibliography{ref}






\newpage
\section{A. Extensive Proof for Section Face alignment}
 Denote $\mathbf{I}^a_i$ as the transferred image based on landmarks $R_i$.  We seek an optimized affine transformation matrix $\mathbf{T}_i^*$ to transfer a face image $\mathbf I$ to $\mathbf{I}^a_i$. It can be proved:
\begin{equation}
\label{eq:cal_matrix}
\begin{aligned}
    \mathbf{T}_i^* & =  \text{argmin}_\mathbf{T} \| \mathbf{T}[ldmks(\mathbf I)] - R_i\|_2 \\
    & =  \text{argmin}_\mathbf{T} \| \mathbf{T}[ldmks(\mathbf I)] - \mathbf{A}_iR_0 \|_2 \\
    & =  \text{argmin}_\mathbf{T} \| \mathbf{A}_i^{-1}\mathbf{T}[ldmks(\mathbf I)] - R_0 \|_2 \\
    & =   \mathbf{A}_i\mathbf{T}_0^*
\end{aligned}
\end{equation}

\section{B. Datasets Details}
The number of identities and images of datasets in the benchmark are listed in Table \ref{table:datasets}.

\begin{table}[H]
\begin{center}
\small
\begin{tabular}{c|c|c}
\hline
Datasets & \; \#Identity & \; \#Image \\
\hline
 CASIA & 10575 & 0.49M \\ [0.5ex]
 MS-Celeb-1M-v1c & 86,876 & 3.92M \\
 LFW & 5,749 & 13,233 \\
 CALFW & 5,749 & 12,174  \\
 CPLFW & 5,749 & 11,652 \\
 AgeDB-30 & 568 & 16,488 \\
 MultiPIE & 337 & 0.75M \\
 IJB-A & 500 & 25,813 \\
 CCW & 6,317 & 41,373 \\
 \textit{Reduced MS-Celeb-1M-v1c} & 30,000 & 0.90M \\
 \hline 
\end{tabular}
\caption{Face datasets of FAPS benchmark.}
\label{table:datasets}
\end{center}
\end{table}
For a fair comparison, we choose datasets with intact face images for the benchmark. CFP-FP is not included due to heavily truncated images. LFW is collected in unconstrained environments with high color jittering and illumination variations. AgeDB-30's primary difficulty lies in the large age gaps and CPLFW has large face pose variations. CALFW demonstrates the age challenge in the wild. MultiPIE is a large multi-view face recognition benchmark.
It splits the test data on different yaw angles. We test our model on subsets of $\pm 90^{\circ}, \pm 75^{\circ}, \pm 60^{\circ}$ yaw angles to evaluate the performance in a large pose situation. The protocol from \cite{Zhou2018} are followed, where the last 137 subjects with 13 poses, 20 illuminations and neutral expression are selected for testing.
Euler angle distributions of full MS-Celeb-1M-v1c and \textit{Reduced MS-Celeb-1M-v1c} are enforced to be consistent to reduce the gaps between two datasets. We also test the proposed FAPS on IJB-A after training on the large scale dataset MS-Celeb-1M-v1c. Compared with previous datasets, the faces in IJB-A have larger variations and present a more unconstrained scenario.

\section{C. Setting Details}
Due to the length constraint, we list the some settings here:
\begin{itemize}
    \item During training, the weight decay is set to 0.0005 and the momentum is 0.9.
    \item All image pixel values are subtracted with the mean 127.5 and divided by 128.
    \item During training, horizontally flipping with probability 0.5 are used as the data augmentation.
    \item During searching, the weight decay is set to 0.0015. The statistics of all the Batch Normalization (BN) operations are recalculated once the alignment policy changed.
    \item Cosine annealing learning rate that decays from 0.1 to 0.00001 is applied as LR-scheduler to smooth the process.
    \item The momentum is set to 0 to eliminate the impact of the input changes.
    \item The random seeds are all set to ``1234" during searching and training process.
    \item During testing, for a straightforward comparison on MultiPIE, no further training or finetuning on the first 200 subjects are conducted.
\end{itemize}

\section{D. Generalization of the searched policy}
We evaluate the performance consistency between ResNet18 and ResNet50. Since searching and training networks differs, it's essential to ensure the generalization of the searched policy.
We train ResNet18 on CASIA with the same settings as ResNet50. Table \ref{table:casia_r18} shows the results.
Our searched alignment policy outperforms the compared methods at all datasets, much like the results of ResNet50 in Table \ref{table:casia_r50_1}.  It reveals that searching with ResNet18 won't introduce a performance gap.

\begin{table}[H]
\begin{center}
\small
\begin{tabular}{c|c|c|c|c}
\hline
Alignment Policy & LFW &  AgeDB-30 & CALFW & CPLFW \\
\hline
ArcFace (190,-7)  & 99.10  & 93.18 & 89.05 & 78.43 \\
MFR (198,-15)  & 99.12 & 93.30 & 89.45 & 79.22 \\
\hline
\textit{TigthROI} (160,0) & 99.02 & 93.73  & 88.78 &  79.30  \\
\textit{SuperROI} (232,0) & 99.18 & 93.38 & 88.80 & 79.22 \\
\hline
FAPS$_C$ (192,4) & \textbf{99.20} & \textbf{94.02} & \textbf{89.47} & \textbf{80.28} \\
\hline
\end{tabular}
\caption{\normalsize{Verification performance  (\%) with different alignment policies on ResNet18.}}
\label{table:casia_r18}
\end{center}
\end{table}



\end{document}